%% file: main.tex
\documentclass[letterpaper, 10 pt, conference]{ieeeconf}  

\IEEEoverridecommandlockouts                              

\usepackage[T1]{fontenc}

\usepackage{amsfonts}	
\usepackage{amsmath}	
\usepackage{amssymb}    
\usepackage{siunitx}
\usepackage{xfrac}    
\usepackage{pifont}   

\usepackage{booktabs}
\usepackage{makecell}  
\usepackage[flushleft]{threeparttable}  
\usepackage{multirow}
\usepackage{rotating}

\usepackage{xspace}    
\usepackage[usenames,dvipsnames]{xcolor}    
\usepackage{colortbl}
\let\labelindent\relax
\usepackage[inline]{enumitem} 
\usepackage{graphicx} 
\usepackage{microtype}
\usepackage{cite}
\usepackage{flushend}
\usepackage{xcolor}
\usepackage{placeins}
\usepackage{algorithm}
\usepackage{algpseudocode}
\algrenewcommand\algorithmiccomment[1]{\hfill\(\triangleright\) #1}

\makeatletter
\let\NAT@parse\undefined
\makeatother

\usepackage{url}

\usepackage[pdfencoding=auto, colorlinks=true]{hyperref} 
\usepackage[hang,flushmargin]{footmisc}

\usepackage[capitalize]{cleveref}
\crefname{section}{Sec.}{Secs.}
\Crefname{section}{Section}{Sections}
\Crefname{table}{Table}{Tables}
\crefname{table}{Tab.}{Tabs.}

\definecolor{darkgreen}{rgb}{0.0, 0.8, 0.0}
\definecolor{darkred}{rgb}{0.8, 0.0, 0.0}

\begin{document}

\newcommand{\myfunding}{This work has been supported by project PID2024-161576OB-I00, funded by MCIN/AEI/10.13039/501100011033 and co-funded by the European Regional Development Fund (ERDF, “A way of making Europe”), by project PLEC2023-010343 (INARTRANS 4.0) funded by MCIN/AEI/10.13039/501100011033, by the R\&D program TEC-2024/TEC-62 (iRoboCity2030-CM) and ELLIS Unit Madrid, granted by the Community of Madrid, and Spanish MICIU through a FPU grant.}

\title{\LARGE \bf
TGRIP: A Text-Guided Approach to Vehicle Instance Prediction in Autonomous Driving
}



\author{
Miguel Antunes-García$^{1}$,
Santiago Montiel-Marín$^{1}$,
Fabio Sánchez-García$^{1}$, \\
Rodrigo Gutiérrez-Moreno$^{1}$,
Rafael Barea$^{1}$,
and Luis M. Bergasa$^{1}$
\thanks{$^{1}$ Electronics Department, University of Alcalá (UAH), Alcalá de Henares, Spain.}%
\thanks{\myfunding}%
}

\maketitle
\thispagestyle{empty}
\pagestyle{empty}


\begin{abstract}
    \input{sections/0_abstract}
\end{abstract}


\input{sections/1_introduction}
\input{sections/2_related_work}
\input{sections/3_method}
\input{sections/4_experiments}
\input{sections/5_conclusion}


\FloatBarrier

\footnotesize
\bibliographystyle{IEEEtran}
\bibliography{biblio}


\end{document}

%% file: sections/0_abstract.tex
Bird's-Eye View (BEV) end-to-end instance prediction has emerged as a robust paradigm for autonomous driving perception, effectively mitigating the error propagation inherent in traditional modular pipelines.
However, current state-of-the-art approaches rely predominantly on geometric supervision, such as occupancy regression and optical flow, effectively treating scene agents as generic moving obstacles.
This absence of explicit semantic awareness imposes limitations on the capacity of the model to solve ambiguities in complex scenarios, particularly those where object-specific behavior is essential for accurate forecasting (e.g. overtaking, intersections).
In this paper, we introduce \textbf{T}ext-\textbf{G}uided \textbf{R}epresentation for \textbf{I}nstance \textbf{P}rediction (TGRIP), a novel framework that bridges this gap by injecting rich semantic priors into the instance prediction loop.
The proposed teacher-student pipeline employs Vision-Language Foundation Models to generate dense, semantic-enhanced BEV maps from multi-camera images. These maps serve as auxiliary supervision during training, guiding the network to learn spatio-temporal representations that are not only geometrically consistent but also semantically discriminative.
To the best of our knowledge, this represents the first attempt to unify semantic guidance with the temporal task of future instance prediction.
The experimental results demonstrate that TGRIP surpasses existing state-of-the-art models in nuScenes, validating the hypothesis that semantic enrichment is a fundamental element for robust, end-to-end motion prediction.
Code is available on \href{https://github.com/miguelag99/TGRIP}{https://github.com/miguelag99/TGRIP}.

%% file: sections/1_introduction.tex
\section{Introduction}
\label{sec:intro}

Accurate object prediction is fundamental for ensuring safety in autonomous driving systems.
In order to successfully navigate complex dynamic environments, an autonomous vehicle must be capable of not only perceiving its surroundings, but also anticipating the future trajectories of surrounding agents.
Conventional modular pipelines address this by sequentially executing detection, tracking, and trajectory prediction tasks \cite{casas2018intentnet, chai2019multipath, gomez2023efficient}. However, these systems frequently exhibit compounding errors, where a missed detection or a switched ID in the early stages results in catastrophic prediction failures downstream.

\begin{figure*}[t]
  \centering
  \includegraphics[width=0.7\linewidth]{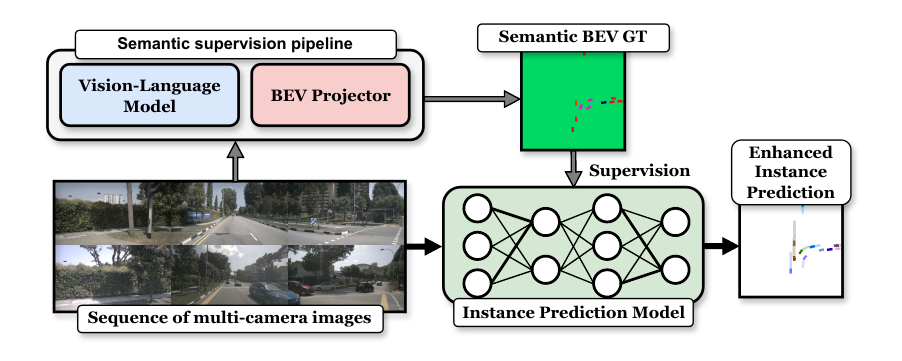}
  \caption{TGRIP incorporates supplementary semantic supervision only during the training process by leveraging a foundational Vision-Language model to generate semantic-enhanced BEV ground truth.}
  \label{fig:teaser}
\end{figure*}

In order to address this limitation, the field has evolved to focus on 360º End-to-End Instance Prediction \cite{fiery2021, powerbev, fast_effic}. These approaches map raw sensor data from a multi-camera setup directly to a dense Bird's-Eye View (BEV) representation, predicting future states through dense occupancy and flow grids rather than explicit trajectory coordinates.
While these methods have been shown to achieve remarkable results by leveraging this holistic scene representation, they predominantly rely on geometric descriptors, such as occupancy and optical flow consistency, to propagate instance states.
These methods excel at predicting where generic pixels might move but often lack a deep understanding of what those pixels represent.
Simultaneously, the emergence of Vision-Language Foundation Models (e.g., CLIP \cite{clip}, SAM \cite{sam}) has illustrated that visual representations can be significantly enriched through open-vocabulary semantic alignment.
However, the vast majority of research in this area has focused on 2D image or static 3D tasks \cite{vobecky2023pop3d}, thereby leaving the temporal domain of motion prediction largely unexplored.

In this work, we argue that semantic understanding is not merely an additional task, but rather a fundamental element that enables robust motion prediction. We introduce \textbf{T}ext-\textbf{G}uided \textbf{R}epresentation for \textbf{I}nstance \textbf{P}rediction (TGRIP), a novel framework that injects dense semantic supervision into the end-to-end instance prediction loop.
In contrast to prior methods that depend exclusively on geometric ground truth, our approach employs a teacher pipeline, show in Figure \ref{fig:teaser}, driven by multi-camera foundation models to generate semantic-enhanced BEV maps.
These maps act as a supplementary supervisory signal during training, forcing the network to learn instance-specific features that are both geometrically consistent and semantically meaningful.
By explicitly guiding the model to recognize the semantic identity of scene elements, TGRIP enhances its capacity to discern closely, interacting instances and predict their motion patterns.
Experimental results demonstrate that this semantic injection leads to superior instance association and trajectory forecasting, surpassing state-of-the-art performance on nuScenes \cite{nuscenes2019}.
To summarize, our main contributions are as follows:
\begin{enumerate}
    \item We present a \textbf{novel pipeline} for generating high fidelity semantic BEV maps from multi-camera images, offering a comprehensive source of auxiliary supervision that surpasses conventional labels.
    \item To the best of our knowledge, TGRIP represents the first attempt to \textbf{unify vocabulary semantic guidance with} the temporal task of \textbf{end-to-end instance prediction}.
    \item The semantic supervision is integrated into a robust baseline, demonstrating that \textbf{TGRIP surpasses existing state-of-the-art instance prediction models}.
\end{enumerate}

The findings serve to corroborate our fundamental hypothesis, which states that the incorporation of semantic priors into the BEV representation is critical for the identification and prediction of main actors in autonomous driving maneuvers within dynamic scenes.

%% file: sections/2_related_work.tex
\section{State of the Art}
\subsection{Vision-Language Learning Foundations}
\label{subsec:sota-vl}

Recent advancements in large scale representation learning have led to a paradigm shift from closed-set supervised training to a new approach involving open-world foundation models.
The utilization of substantial web-scale datasets by these architectures facilitates the acquisition of robust, transferable representations, effectively mitigating the disparity between visual perception and semantic understanding.
The pivotal aspect of this transformation involves the integration of image and text modalities into a unified feature space.
CLIP \cite{clip}, a pioneering approach in this field, utilizes contrastive learning on image-text pairs, thereby demonstrating zero-shot classification and retrieval capabilities that are typically absent in traditional supervised models.
In a similar line of research, ALIGN \cite{align} demonstrates that the scale of the training data can mitigate the impact of noise, leveraging over one billion noisy image alt-text pairs to achieve state-of-the-art representations without the need for expensive filtering or post-processing.
However, the standard contrastive loss objective imposes the necessity of costly global batch synchronization in order to normalize the similarities across the batch.
In response to these scalability limitations, SigLIP \cite{siglip} introduced a pairwise sigmoid loss, effectively decoupling batch size from loss definition.
SigLIP 2 \cite{siglip2} builds directly on top of this architecture, addressing the limitations of standard contrastive models in fine-grained localization and dense prediction by employing a unified training recipe.
While global alignment models demonstrate effectiveness in image-level understanding, dense prediction tasks require fine-grained geometric and local semantic features.
In order to address this disparity, MaskCLIP \cite{maskclip} adapts the pre-trained CLIP image encoder by modifying the final attention pooling layer to extract dense, patch-level features rather than a global embedding.
This modification enables annotation-free, open-vocabulary semantic segmentation.
In a similar manner, ZegCLIP \cite{zegclip} extends CLIP to pixel-level tasks while addressing the overfitting issues of simple text-patch matching.
It achieves this by introducing a "Relationship Descriptor" that incorporates image-level priors into text embeddings and using a non-mutually exclusive loss. This combination significantly improves generalization to unseen classes in a one-stage, efficient framework.
Concurrently, self-supervised approaches like the DINO series \cite{dino, dinov2, dinov3} leverage Vision Transformers (ViT) to generate potent, localized semantic descriptors without explicit labels, while the Segment Anything Model (SAM) models \cite{sam, sam2, sam3} have established a paradigm for precise, class-agnostic segmentation masks.
These "universal" vision models complement the semantic scope of CLIP-style models by supplying the necessary geometric precision required for complex scene understanding and object discrimination.
Recent advancements in the field of Vision-Language Models (VLMs) have led to the integration of Large Language Models (LLMs) for complex reasoning over visual data.
Although early generalist models, such as LLaVA \cite{llava}, were among the first to explore this idea, recent advancements, including Qwen3-VL \cite{li2026qwen3} and InternVL \cite{internvl25, zhu2025internvl3}, have led to substantial improvements in performance on high-resolution inputs and fine-grained spatial grounding, critical in complex scenarios.
While these models represent the upper limit of semantic reasoning, they frequently demand substantial computational resources for real-time inference.

For the purposes of this study, it is essential to interpret these foundation models not merely as inference components, but rather as rich sources of information suitable for the supervision of other architectures.
In this paper, we leverage the aligned feature space of CLIP and SigLIP2 to provide auxiliary semantic supervision during the training process of TGRIP.
By ensuring semantic consistency between our BEV features and CLIP/SigLIP2 embedding space, we introduce discriminative prior information that guides the network to learn more robust instance representations. This, in turn, enhances prediction performance beyond what is possible with geometric labels alone.

\subsection{Semantic Representation in Autonomous Driving}
\label{subsec:sota-semantic-rep}

The success of foundation models has led to a paradigm shift in the way autonomous driving perception architectures are designed. These architectures have transitioned from closed-set, predefined object categories (e.g. vehicles, pedestrians), to open-vocabulary scene understanding and semantic infusion.

Recent 2D semantic-guided approaches, such as OpenWorldSAM \cite{xiao2025openworldsam} and LPOSS \cite{stojnic2025lposs}, have demonstrated the effectiveness of propagating semantic labels across images to identify complex objects.
However, for autonomous driving applications, these 2D priors must be converted into a unified perspective in 3D or Bird's-Eye View (BEV).
Addressing this spatial transition is a core challenge that has driven recent breakthroughs in semantic occupancy and open-vocabulary mapping.
An approach to addressing this challenge involves distilling knowledge from vision-language models into 3D representations.
Methods such as TPVFormer \cite{tvpformer} and OccFormer \cite{zhang2023occformer} focus on dense semantic 3D occupancy prediction, while subsequent works, such as POP-3D \cite{vobecky2023pop3d} and OVO \cite{tan2023ovo} explicitly produce semantic-rich voxel maps, suitable for open-vocabulary retrieval.
These approaches generally adopt a teacher-student paradigm, in which 2D foundation models, similar to the ones discussed in section \ref{subsec:sota-vl}, generate pseudo-labels or dense feature maps to supervise the final 3D network.
Although full 3D methods offer a comprehensive and detailed representation of the scene, they often require high computational requirements to work.
Other architectures such as Talk2BEV \cite{choudhary2024talk2bev}, focus on integrating semantic capabilities in a BEV representation through a training-free framework that interconnects Large Vision-Language Models (LVLMs) with standard grid-based maps.
Rather than relying on closed-set perception, Talk2BEV constructs a "language-enhanced map" by projecting BEV objects back onto the camera images and employing LVLMs to generate detailed semantic captions.
This text-based JSON representation integrates geometric cues with semantic descriptions, enabling a general-purpose LLM to perform complex visual reasoning.

Despite these advancements, the integration of explicit semantic information remains fragmented across different spatial representations.
While recent research in 3D Semantic Occupancy has achieved high-fidelity voxel-based classification, these methods are often limited to static scene reconstruction at the current timestamp. This limitation incurs in high computational costs that compromise long-term temporal reasoning.
BEV representations have become the prevailing standard for efficient autonomous driving perception \cite{li2024bevformer, peng2023bevsegformer, gaussiancar, chambon2024pointbev, fbbev} and planning \cite{bevplanner, shao2024lmdrive, winter2025bevdriver, internfuser, shao2023reasonnet}, frequently exhibiting a deficiency in the dense semantic richness characteristic of open-vocabulary 3D voxel grids.
The infusion of dense, aligned semantic priors into temporal BEV representations has not yet been thoroughly explored for temporal tasks.
TGRIP addresses this limitation by introducing a novel semantic supervision framework that enhances the discriminative power of the BEV feature space.
Contrary to prior semantic-aware methods, which were constrained to static scene parsing \cite{choudhary2024talk2bev} or voxel classification \cite{vobecky2023pop3d,tan2023ovo}, TGRIP employs this BEV semantic depth to enhance the efficacy of end-to-end instance prediction.

\begin{figure*}[h]
  \centering
  \includegraphics[width=0.99\linewidth]{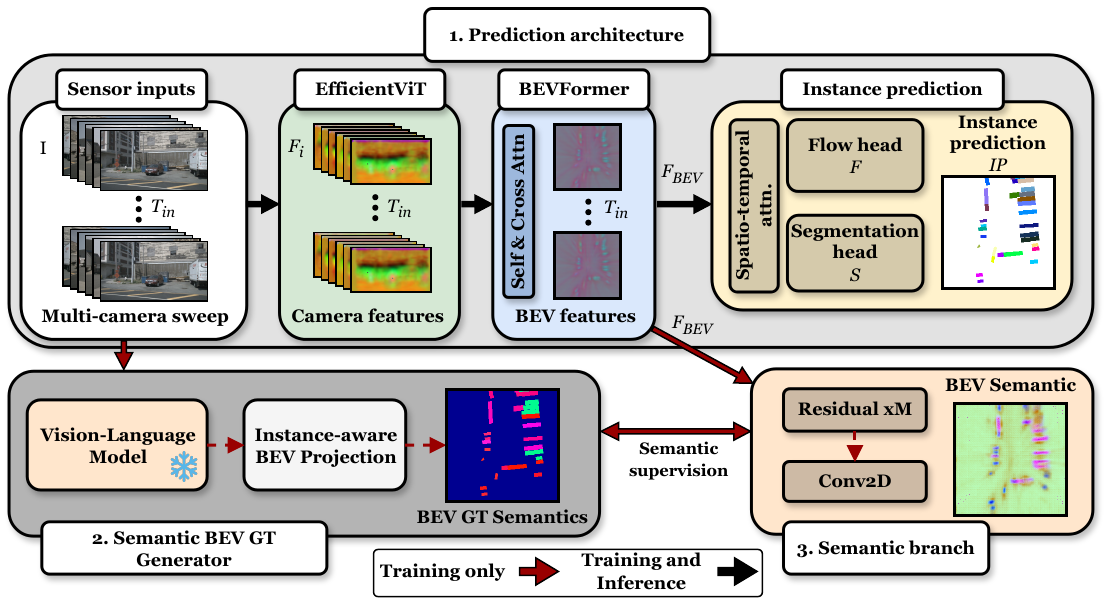}
  \caption{Overview of the TGRIP framework. A specialized pipeline is employed to generate semantic BEV ground truth, which is then used to supervise an auxiliary semantic head during the training process.}
  \label{fig:architecture}
\end{figure*}

\subsection{Vehicle Prediction in Autonomous Driving}
\label{subsec:sota-prediction}

Classical approaches of vehicle motion prediction normally are part of a multi-stage perception architecture, where the history of each agent is obtained trough a detection and tracking stage before extracting the future trajectories in the prediction module.
The past and future information is processed in a vectorized manner, where the positions and trajectories are characterized by the direct coordinates in the scene.
Methods such as IntentNet \cite{casas2018intentnet} or Multipath \cite{chai2019multipath} rely on Convolutional Neural Networks (CNNs) to extract the scene features.
Due to the loss of details of these approaches, VectorNet \cite{gao2020vectornet} and LaneGCN \cite{lanegcn} leveraged a graph-based encoders to better preserve details and interactions between actors.
To improve long-term interaction, methods such as Scene Transformer \cite{ngiam2021scene} or HiVT \cite{zhou2022hivt} switch from graph-based processing to attention mechanisms present in Transformers, capturing contexts more efficiently from the input data.
The different decoding strategies in these motion prediction architectures have also evolved to incorporate new techniques.
MTR \cite{mtr} uses "motion query pairs" to combine global intention localization with local refinement. QCNet \cite{qcenet} uses a two-stage, anchor-free proposal system.
These query-centric approaches capture multi-modal intentions and social interactions more flexibly than rigid, anchor-based constraints.
Despite the complexity of these vectorized architectures, they are still susceptible to error propagation from the upstream perception pipeline, which can result in amplified noise, misdetections, or association errors during the initial stages, leading to unrealistic or failed trajectory forecasts.
This dependency introduces a bottleneck in which the final motion prediction is strictly constrained by the accuracy of the preceding discrete stages. Consequently, there is a motivation to shift toward end-to-end architectures that reason directly from raw sensor data.

Recent research has shifted toward end-to-end architectures that reason directly from raw sensor data to mitigate the error propagation inherent in modular pipelines.
VIP3D \cite{gu2023vip3d} and DeTra \cite{casas2024detra} are notable examples of methods that unify perception and prediction, decoding the scene into sparse trajectory sets, similar to modular motion prediction.
While trajectory-based approaches have proven effective for certain agents, they have been observed to encounter challenges in modeling holistic scene dynamics and dense interactions when compared to BEV grid-based alternatives.
In contrast, End-to-End Instance Prediction models operate exclusively within a unified BEV representation, predicting future states via dense segmentation and flow maps rather than explicit coordinates.
FIERY \cite{fiery2021} was a pioneer in this dense approach, while BEVerse \cite{beverse} integrated detection, mapping, and motion into a single efficient pipeline.
In order to address the issue of long-term blurring, StretchBEV \cite{stretchbev} introduced stochastic residual updates, and methods such as Fast and Efficient \cite{fast_effic} optimized these dense representations for real-time inference.
PowerBEV \cite{powerbev} simplifies complex post-processing by only requiring flow and segmentation maps, while S3-P3 \cite{hu2022stp3} incorporates an instance prediction stage inside an end-to-end driving model.
DMP \cite{dmp} introduces a difference-guided block that enhances long temporal understanding, while BEVPredFormer \cite{bevpredformer} introduces a temporal attention module that further improves temporal and spatial reasoning capabilities.

Nevertheless, a significant constraint remains: contemporary grid methods primarily depend on geometric supervision, lacking explicit semantic awareness of the objects they predict.
TGRIP addresses this limitation by incorporating vocabulary semantic knowledge into a robust BEV end-to-end prediction pipeline, thereby demonstrating the importance of semantic consistency in enhancing instance separation and long-term motion forecasting.

%% file: sections/3_method.tex
\begin{figure*}[h]
  \centering
  \includegraphics[width=0.99\linewidth]{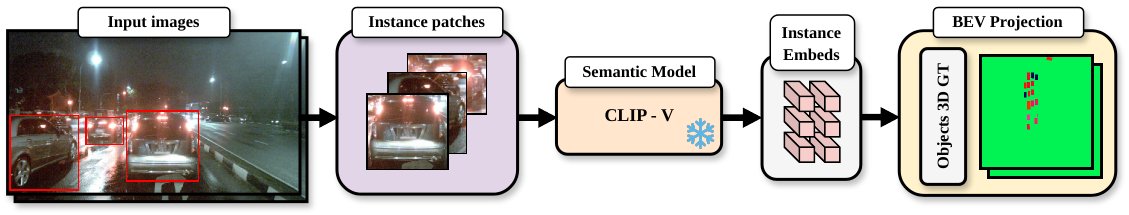}
  \caption{Overview of the BEV semantic ground truth generation pipeline.
  Object-level crops are first extracted from the input images using the 3D ground truth annotations provided by the dataset. Subsequently, each crop is processed by a semantic model to generate a per-instance embedding. Finally, these embeddings are projected into the BEV space using the provided geometric ground truth.}
  \label{fig:bev_semantic_gt_generator}
\end{figure*}

\section{Methodology}
\label{sec:method}

As shown in Figure \ref{fig:architecture}, TGRIP is organized into three primary components:
1) A baseline instance prediction network that independently generates the necessary information required for instance prediction. 2) A semantic ground truth generator that constructs the semantic-enhanced BEV maps employed for supervision. 3) An auxiliary semantic branch that is appended to the base network during training to implement the enhanced semantic supervision.

\subsection{Prediction Architecture}
\label{subsec:arch}

The input data for the instance prediction task consists of a sequence of $T_{in}$ multi-camera frames, $I \in \mathbb{R}^{T_{in} \times N_c \times C_{im} \times H_{im} \times W_{im}}$, incorporating both historical and present observations.
In order to account for camera geometry in the BEV projection stage, the framework includes intrinsic parameters $K \in \mathbb{R}^{N_c \times 3 \times 3}$ and extrinsic transformations $R \in \mathbb{R}^{N_c \times 4 \times 4}$ for the $N_c$ cameras on the ego-vehicle.
Additionally, previous ego-positions $P$ are integrated to compensate for ego-motion across the input sequence.
The model outputs include a future segmentation map $S \in \mathbb{R}^{T_{out} \times N_c \times H \times W}$ and a future flow map $F \in \mathbb{R}^{T_{out} \times 2 \times H \times W}$.
The primary objective is the generation of a predicted BEV instance map, $IP \in \mathbb{R}^{H \times W}$, in a post-processing step, where each predicted instance is represented with a different id and color.

The TGRIP architecture first extracts multi-scale features from the multi-camera input sweep. To achieve high-efficiency feature extraction, the model employs EfficientViT \cite{liu2023efficientvit} as its backbone.
A lightweight neck fuses the multiple features of each camera into a fused feature map $F_i$.
To map these camera-plane features into a unified coordinate system, the architecture employs a spatial transformation module. Specifically, TGRIP employs the attention-based mechanisms introduced in BEVFormer \cite{li2024bevformer}, utilizing both self-attention and cross-attention modules to project image-space features into a common BEV representation, obtaining $T_{in}$ BEV feature maps $F_{BEV} \in \mathbb{R}^{T_{in} \times C \times H \times W}$.

Subsequent to feature extraction, the model executes the final instance prediction.
TGRIP first incorporates the spatio-temporal module proposed in BEVPredFormer \cite{bevpredformer} to better capture the dependencies between grid cells within and between historical frames.
These refined features are subsequently processed by the prediction heads.
Our methodology is guided by established SOTA methodologies \cite{powerbev, fast_effic, dmp}, employing a dual pyramid architecture, comprised by an encoder, predictor, and decoder, to project the features into $T_{out}$ future timestamps and generate the corresponding flow $F$ and segmentation $S$ maps.
In a final post-processing step, the flow information is used to propagate the instances extracted from the segmentation map into $T_{out}$, obtaining the final prediction map $IP$.

\subsection{Semantic Ground Truth Generator}
\label{subsec:semantic_gt}

Standard autonomous driving datasets generally lack the complete ground truth necessary for the proposed BEV semantic supervision. To address this limitation, a specialized pipeline, summarized in Algorithm \ref{alg:bev_semantic_gt}, was developed to generate the required BEV semantic maps.
As shown in Figure \ref{fig:bev_semantic_gt_generator}, this process faces two primary challenges. First, it involves the extraction of high-level semantic features for each instance from the image plane. Second, it involves accurate projection of these features into a BEV that is spatially aligned with the primary instance prediction task.

To obtain visual features inherently aligned with high-level semantic information, we use the pretrained CLIP \cite{clip} latent space to extract instance-specific embeddings.
The process begins with the projection of the 3D object detection labels provided by the dataset onto the corresponding camera planes for each instance.
If an object is visible across multiple cameras, we select the 2D bounding box that maximizes the object's visibility. Objects that are completely occluded are excluded from processing.
These refined 2D bounding boxes are used to crop the regions of interest (ROIs) from the source images. The CLIP image encoder then processes these crops to generate high-dimensional object embeddings $E_{obj} \in \mathbb{R}^{ C \times N_{instances}}$ for each corresponding timestamp.

Given that the image crops are derived directly from the original object labels, an explicit spatial correspondence between each semantic embedding and its respective 3D object is maintained.
The generation of the final semantic-enhanced BEV ground truth $V \in \mathbb{R}^{ C_{CLIP} \times H \times W}$ is achieved through the application of the same geometrical projection method that is employed to generate the flow and segmentation ground truth maps.
Specifically, for each grid cell occupied by an instance in the BEV plane, the feature vector is filled with the corresponding CLIP-based embedding extracted from the image crops.

\begin{algorithm}[h]
\caption{BEV Semantic Ground Truth Generation}
\label{alg:bev_semantic_gt}
\begin{algorithmic}[1]
\Require 3D bounding box labels per instance $\mathcal{L}$, multi-camera source images  $\mathcal{I}$, camera intrinsic and extrinsic calibration $\mathcal{K}$, BEV grid $\mathcal{G}$ with spatial resolution $H \times W$
\Ensure semantic-enhanced BEV ground truth map  $V \in \mathbb{R}^{C_{\mathrm{CLIP}} \times H \times W}$

\State $V \leftarrow \mathbf{0}^{C_{\mathrm{CLIP}} \times H \times W}$
\State $E_{\mathrm{obj}} \leftarrow \mathbf{0}^{C \times N_{\mathrm{instances}}}$

\For{each instance $i$ in $\mathcal{L}$}
    \State $\mathcal{B}_{i} \leftarrow \emptyset$
    \For{each camera $c$ in $\mathcal{C}$}
        \State $b_{i,c} \leftarrow \mathtt{Project3D2D}(\mathcal{L}_{i},\, \mathcal{K}_{c})$
        \If{$b_{i,c}$ is visible}
            \State $\mathcal{B}_{i} \leftarrow \mathcal{B}_{i} \cup \{b_{i,c}\}$
        \EndIf
    \EndFor
    \If{$\mathcal{B}_{i} = \emptyset$}
        \State \textbf{continue}
    \EndIf
    \State $b_{i}^{*} \leftarrow \arg\max_{b \in \mathcal{B}_{i}}\; \mathtt{Visibility}(b)$
    \State $\tilde{I}_{i} \leftarrow \mathtt{ExtractROI}(\mathcal{I},\, b_{i}^{*})$
    \State $e_{i} \leftarrow \mathtt{CLIP}_{\phi}(\tilde{I}_{i}), \quad e_{i} \in \mathbb{R}^{C_{\mathrm{CLIP}}}$
    \State $E_{\mathrm{obj}}[\,:,\, i] \leftarrow e_{i}$
    \State $\mathcal{G}_{i} \leftarrow \mathtt{BEVCells}(\mathcal{L}_{i},\, \mathcal{G})$
    \For{each cell $(u, v)$ in $\mathcal{G}_{i}$}
        \State $V[\,:,\, u,\, v] \leftarrow e_{i}$
    \EndFor
\EndFor

\State \Return $V$
\end{algorithmic}
\end{algorithm}

\subsection{Semantic Branch}
\label{subsec:semantic_branch}

The primary objective of this module is to distill the aligned semantic knowledge of vision-language models into the BEV representation.
In contrast to conventional heads that are designed to predict discrete categories (e.g., segmentation), this component is intended to generate a per-cell semantic embedding that maintains cross-modal alignment with the scene's elements.
As detailed in Figure \ref{fig:architecture}, this head acts principally as an auxiliary supervision signal during the training process.
By levering the latent BEV features common to all prediction heads $F_{BEV}$, the auxiliary task ensures that rich semantic information is propagated throughout the shared network backbone.
This joint supervision contributes to the refinement of the unified feature representation during training, thereby enhancing the predictive capacity of the model, even in the absence of the semantic head during inference.

Given its role as an auxiliary component, the semantic head is designed to avoid introducing excessive architectural overhead.
In order to evade computationally demanding three-dimensional operations, the module initially performs temporal aggregation by collapsing the temporal and channel dimensions of the input BEV features, $F_{BEV} \in \mathbb{R}^{(T_{in}*C) \times H \times W}$.
A convolutional layer then fuses these flattened maps.
The resulting features are processed through $M$ residual blocks, each comprising a sequence of \texttt{Conv2D, BN, ReLU, Dropout, Conv2D,} and \texttt{BN} layers. Finally, a projection layer maps the features into the desired $C_{CLIP}$-dimensional latent space, obtaining the output $BEV_{semantic} \in \mathbb{R}^{C_{CLIP} \times H \times W}$.

\subsection{Losses}
\label{subsec:losses}

The model is optimized using a two-stage training strategy. First, the baseline instance prediction model is trained without auxiliary semantic supervision.
For this task, we use a multi-task loss function with three components:
\begin{itemize}
    \item \textbf{Smooth L1 loss} for flow estimation ($L_{flow}$) with threshold $\beta = 1.0$, computed exclusively over occupied BEV cells.
    \item \textbf{CrossEntropy loss} for segmentation ($L_{seg}$). To address the foreground-background class imbalance, we assign twice the weight to the vehicle class. Additionally, since future BEV frames are inherently more uncertain, we apply a temporal discount factor $\gamma_t \in (0, 1]$ that decreases with the future timestep $t$.
    \item \textbf{L2 loss} for auxiliary centerness supervision ($L_{cntr}$) generated by placing a 2D Gaussian kernel centered at each BEV centroid, producing a soft heatmap.
\end{itemize}
We utilize a dynamic weighting strategy that updates throughout the training process to balance these objectives.

During the second training stage, we integrate the auxiliary semantic branch into the architecture.
We initialize this stage with the pre-trained weights from the first phase and use a reduced learning rate to effectively merge the existing knowledge with the new semantic information.
We optimize the semantic head using \textbf{Cosine Similarity loss} ($L_{sem}$), which measures the alignment between the predicted and ground-truth embeddings for the whole BEV map.
During this phase, the entire model undergoes fine-tuning under full supervision according to the joint loss function defined in Equation \ref{eq:losses}, with loss contributions balanced via dynamic weighting, similar to the first stage.

\begin{equation}
  \mathcal{L=}\lambda_1 \mathcal{L}_{flow} + \lambda_2 \mathcal{L}_{seg} + \lambda_3 \mathcal{L}_{cntr} + \lambda_4 \mathcal{L}_{sem}
  \label{eq:losses}
\end{equation}

%% file: sections/4_experiments.tex
\begin{table*}[h]
\centering
\caption{Instance prediction performance on nuScenes validation set. All results are obtained with the official implementation if available.}
\label{tab:results}
\setlength{\tabcolsep}{6pt}
\begin{tabular}{@{}cccccccc@{}}
\toprule
\multirow{2}{*}{\textbf{Model}} & \multirow{2}{*}{\textbf{Code}} & \multirow{2}{*}{\textbf{Semantic Supervision}} & \multirow{2}{*}{\textbf{Semantic Teacher}} & \multicolumn{2}{c}{\textbf{Long range}} & \multicolumn{2}{c}{\textbf{Short range}} \\
                               &                                 &                                                &                                          & \textbf{IoU $\uparrow$}  & \textbf{VPQ $\uparrow$} & \textbf{IoU $\uparrow$} &\textbf{ VPQ $\uparrow$} \\ \midrule
StretchBEV \cite{stretchbev}  & \ding{51} & \ding{55}       & - & 37.1 & 29.0 & 55.5 & 46.0 \\
FaE \cite{fast_effic}         & \ding{51} & \ding{55}       & - & 37.4 & 29.8 & 59.1 & 53.7 \\
Fiery \cite{fiery2021}        & \ding{51} & \ding{55}       & - & 36.7 & 29.9 & 59.4 & 50.2 \\
ST-P3 \cite{hu2022st}         & \ding{51} & \ding{55}       & - & 38.9 & 32.0 & -    & -    \\
PowerBEV \cite{powerbev}      & \ding{51} & \ding{55}       & - & 38.9 & 32.2 & 62.5 & 55.5 \\
BEVerse \cite{beverse}        & \ding{51} & \ding{55}       & - & 38.7 & 33.3 & 61.4 & 54.3 \\
DMP \cite{dmp}                & \textcolor{darkred}{\ding{55}} & \ding{55}       & - & 38.8 & 34.0 & 62.9 & \textbf{57.5} \\ \midrule
Baseline - BEVPredFormer \cite{bevpredformer} & \ding{51} & \ding{55}       & - & 40.9 & 33.3 & 63.9 & 54.9 \\
\textbf{Ours - TGRIP}         & \ding{51} & \textcolor{darkgreen}{\ding{51}} & CLIP-B/16 & \textbf{41.3} \textcolor{darkgreen}{(+0.4)} & \textbf{34.3} \textcolor{darkgreen}{(+1.0)} & \textbf{64.5} \textcolor{darkgreen}{(+0.6)} & 56.1 \textcolor{darkgreen}{(+1.2)}\\
\textbf{Ours - TGRIP}         & \ding{51} & \textcolor{darkgreen}{\ding{51}} & CLIP-L/14 & \textbf{41.3} \textcolor{darkgreen}{(+0.4)} & \textbf{34.3} \textcolor{darkgreen}{(+1.0)} & \textbf{64.5} \textcolor{darkgreen}{(+0.6)} & 56.3 \textcolor{darkgreen}{(+1.4)} \\ \bottomrule
\end{tabular}
\end{table*}

\section{Experiments}

\subsection{Datasets and Metrics}
\label{subsec:data_metric}

The experimental evaluation follows the official split of the nuScenes dataset \cite{nuscenes2019}: 700 scenes for training, 150 scenes for validation, and 150 scenes for testing.
The dataset provides multi-sensor data sampled at a frequency of 2 Hz.
Following other SOTA methodologies, our training and evaluation protocols focus exclusively on the vehicle supercategory, which includes the following specific classes: \textit{car}, \textit{bus}, \textit{truck}, \textit{construction vehicle}, and \textit{motorcycle}.

The evaluation of instance prediction architectures is based on two metrics.
First, Intersection over Union ($IoU$), defined in Equation \ref{eq:iou}, quantifies the spatial overlap between predicted and ground-truth vehicle masks, thereby assessing the model's ability to segment instances within the BEV plane across both current and future frames.
Secondly, Video Panoptic Quality ($VPQ$), as specified in Equation \ref{eq:vpq}, quantifies the capacity of the model to preserve coherent instance identities over temporal extents. The incorporation of both segmentation accuracy and temporal flow quality into VPQ provides a comprehensive measure of spatio-temporal consistency.

\begin{equation}
    \text{IoU}=\frac{1}{T_\text{pred}} \sum_{t=0}^{T_\text{pred}-1} \frac{\hat{y}_{t}^\text{seg} \cap {y}_{t}^\text{seg}}{{\hat{y}_{t}^\text{seg} \cup {y}_{t}^\text{seg}}}
    \label{eq:iou}
\end{equation}

\begin{equation}
    \text{VPQ} = \sum_{t=0}^{T_\text{pred}-1} \frac{\sum_{(p_t,q_t) \in TP_t} \text{IoU}(p_t,q_t)}{|TP_t| + \frac{1}{2}|FP_t| + \frac{1}{2}|FN_t|}
    \label{eq:vpq}
\end{equation}

\subsection{Implementation Details}
\label{subsec:implementation}

Operating at the native nuScenes sampling rate of 2 Hz, the model incorporates 1.0 second of historical context alongside the current observation to form an input sequence of $T_{in}=3$ frames.
In accordance with established state-of-the-art protocols, a future horizon of 2.0 seconds is predicted, along with the current frame ($T_{pred} = 5$).
To facilitate the initialization of instance identities, the architecture also estimates segmentation and flow for the preceding frame ($T=-1$), resulting in a comprehensive output sequence of $T_{out} = 6$ frames.

The BEV representation is characterized by a fixed grid size of $H\times W = 200\times 200$ and a feature depth of $C=128$.
Given this fixed grid dimensionality, the spatial resolution is determined by the specific extent employed for the prediction task.
The evaluation of the model was conducted across two distinct spatial ranges: a long-range configuration extending 50 meters from the ego-vehicle (covering a $100\text{m} \times 100\text{m}$ area at $0.5\text{m}$ resolution) and a short-range configuration extending 15 meters (covering a $30\text{m} \times 30\text{m}$ area at $0.15\text{m}$ resolution).

The proposed approach employs the EfficientViT-L2 backbone to perform image feature extraction, achieving effective downsampling factors of up to 8 from the original resolution of $448\times800$.
The spatial transformation module integrates six BEVFormer layers, each incorporating temporal self-attention, normalization, spatial cross-attention, and an MLP with a hidden dimension of 1024.
The flow and segmentation prediction heads utilize a five-stage pyramid architecture with channel dimensions of 16, 24, 32, 48, and 64, respectively.
In the context of the auxiliary semantic task, supervision is applied exclusively to the present BEV frame.
The semantic head consists of $M=2$ residual layers, with the final output dimension fixed at $C_{CLIP} = 512$ for CLIP-Base supervision, or $C_{CLIP} = 768$ when utilizing CLIP-Large or SigLIP2.

The model is optimized using AdamW with a One-Cycle learning rate scheduler and an effective batch size of 16.
During the initial pre-training stage (70 epochs), a maximum learning rate of $3e^{-4}$ is adopted. In the subsequent semantic supervision phase, which encompasses 40 epochs, the learning rate is reduced to a maximum of $3e^{-5}$ to ensure stable convergence.
All training and evaluation procedures are conducted on a computational platform equipped with two NVIDIA A100 GPUs.

The teacher pipeline, which incorporates ROI extraction and CLIP encoding, runs entirely offline.
Generating the BEV semantic targets takes approximately three hours for the nuScenes dataset and requires around 1 GB of disk space in FP16 format to avoid data-loading bottlenecks.
Stage 1 of the training process involves training 152M parameters for approximately four days on two NVIDIA A100 GPUs.
Stage 2 involves adding a small semantic head with 39 million parameters, which requires a two-day fine-tuning phase.
Since the semantic head is removed after training, the final deployed model retains the same number of parameters (152 million) and inference time (220 ms on a single A100 GPU) as the baseline model.
This makes the extra training time a favorable trade-off for achieving better predictions without incurring any deployment penalties.

\begin{figure*}[h]
  \centering
  \includegraphics[width=0.99\linewidth]{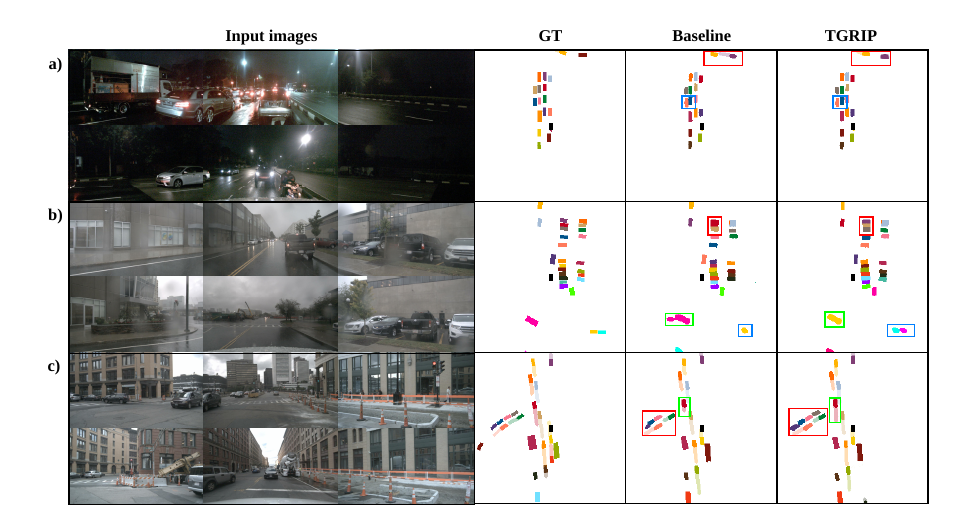}
  \caption{Instance prediction qualitative results in nuScenes of TGRIP and the baseline without semantic supervision. The present frame ego vehicle is always located at the center, represented with a black color. Each detected instance is represented by a different color, using transparency to represent the corresponding future movement.}
  \label{fig:qualitative}
\end{figure*}

\subsection{Quantitative Evaluation}
\label{subsec:quantitative}

\begin{table}[b]
\centering
\caption{Ablation on semantic supervision teaching network used to generate the semantic cues. The TGRIP architecture is fixed in all experiments.}
\label{tab:ablation_teacher}
\setlength{\tabcolsep}{5pt}
\begin{tabular}{@{}ccccc@{}}
\toprule
\multirow{2}{*}{\textbf{Semantic Teacher}} & \multicolumn{2}{c}{\textbf{Long range}} & \multicolumn{2}{c}{\textbf{Short range}} \\
                  & \textbf{IoU $\uparrow$} & \textbf{VPQ $\uparrow$} & \textbf{IoU $\uparrow$} & \textbf{VPQ $\uparrow$} \\ \midrule
\textcolor{darkred}{\ding{55}}  & 40.9 & 33.3 & 63.9 & 54.9 \\
CLIP-B/16   & \textbf{41.3} & \textbf{34.3} & \textbf{64.5} & 56.1 \\
CLIP-B/32   & 41.2 & 34.2 & 64.3 & 55.9 \\
CLIP-L/14   & \textbf{41.3} & \textbf{34.3} & \textbf{64.5} & \textbf{56.3} \\
SigLIP2-B/16& \textbf{41.3} & 34.2 & 64.4 & 56.0 \\ \bottomrule
\end{tabular}
\end{table}

\begin{table}[b]
\centering
\caption{Ablation on the type of semantic information used to perform the supervision. We employ both visual and class information using the corresponding image and text encoders from CLIPB16.}
\label{tab:ablation_cues}
\setlength{\tabcolsep}{5pt}
\begin{tabular}{@{}ccccc@{}}
\toprule
\multirow{2}{*}{\textbf{Semantic Cues}}                            & \multicolumn{2}{c}{\textbf{Long range}}           & \multicolumn{2}{c}{\textbf{Short range}}          \\
                                                                   & \textbf{IoU $\uparrow$} & \textbf{VPQ $\uparrow$} & \textbf{IoU $\uparrow$} & \textbf{VPQ $\uparrow$} \\ \midrule
\textcolor{darkred}{\ding{55}}                                     & 40.9                    & 33.3                    & 63.9                    & 54.9                    \\
Text Class                                                         & 41.1                    & 33.9                    & 64.4                    & 56.0                    \\
Visual                                                             & \textbf{41.3}           & \textbf{34.3}           & \textbf{64.5}           & \textbf{56.3}           \\
Visual + Class                                                     & 41.2                    & 34.2                    & 64.4                    & \textbf{56.3}           \\ \bottomrule
\end{tabular}
\end{table}

A comparative analysis between TGRIP and existing SOTA models on the nuScenes validation split is presented in Table \ref{tab:results}.
The experimental results demonstrate that the incorporation of auxiliary semantic supervision consistently enhances prediction performance across both spatial ranges in comparison with the baseline configuration.
In the long-range setting, TGRIP achieves an IoU of 41.3\% and a VPQ of 34.3\%, representing absolute gains of 0.4\% and 1\%, respectively, over the semantic unsupervised baseline.
Furthermore, TGRIP establishes a new SOTA for IoU in the long-range category, outperforming all existing methods.
In terms of VPQ, our proposal outperforms DMP (which lacks an official implementation) in the long-range setting, achieving a score of 34.3\%. However, DMP exhibits a competitive edge in the short-range scenario, with a VPQ of 57.5\%, closely followed by our own 56.3\%.

To verify the consistency of the performance difference between our method and the baseline, we trained both models three times using different random seeds for the long-range setup.
The baseline model achieved an IoU of $40.85 \pm 0.08$ and a VPQ of $33.23 \pm 0.09$.
Our proposed TGRIP model achieved IoU and VPQ values of $41.28 \pm 0.03$ and $34.26 \pm 0.06$.
The mean and standard deviation for both metrics demonstrate that the improvement over the baseline is clear and not due to randomness in the training process.

Table \ref{tab:ablation_teacher} presents an evaluation of the model's sensitivity to various vision-language architectures employed for semantic distillation.
The experimental results indicate that the integration of any semantic teacher consistently outperforms the baseline configuration across all metrics, thereby confirming the value of cross-modal alignment for instance prediction.
It is worth noting that the observed performance gains are remarkably architecture-agnostic.
Although the larger CLIP-L/14 model achieves the highest overall performance and leads in short-range VPQ over models such as CLIP-B/16 and SigLIP2-B/16, their broader performance remains highly comparable.
In fact, CLIP-B/16 matches the larger model in terms of long-range metrics and short-range IoU.
This finding indicates that the semantic complexity necessary for enhancing BEV instance prediction can be adequately captured even by smaller, more lightweight models.
Consequently, employing architectures like CLIP-B/16 enables a more computationally efficient ground-truth generation process without compromising predictive accuracy.

\begin{figure*}[t]
  \centering
  \includegraphics[width=0.99\linewidth]{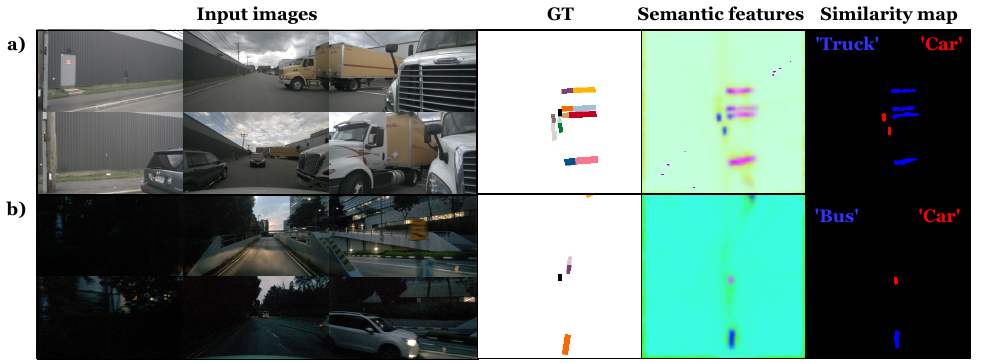}
  \caption{Qualitative visualization of TGRIP semantic maps on nuScenes. The \textit{Semantic features} maps are representations of the high-dimensional output of the semantic head. The \textit{Similarity Map} demonstrates the cosine similarity between these learned BEV features and a target text embedding (e.g., 'car' or 'truck'), thereby illustrating the model's capacity to localize semantic concepts in the BEV plane through vision-language alignment.}
  \label{fig:qualitative2}
\end{figure*}

Table \ref{tab:ablation_cues} investigates the impact of different semantic modalities on the quality of the BEV representation. We compare three distinct supervision strategies: I) Text Class, which uses fixed text embeddings based on category labels; II) Visual, which utilizes per-instance embeddings extracted directly from the CLIP image encoder with the pipeline explained in section \ref{subsec:semantic_gt}; and III) Visual + Class, a fused hybrid approach.
The experimental results demonstrate that visual embeddings provide the most significant performance enhancement.
This finding indicates that the fine-grained, instance-specific information captured by the visual encoder, such as vehicle orientation, scale, and appearance, provides a more comprehensive supervision signal compared to static class-level embeddings.
While the hybrid approach maintains competitive performance, the visual-only cues provide the optimal balance between segmentation accuracy and temporal consistency.

\subsection{Qualitative Results}
\label{subsec:qualitative}

Figure \ref{fig:qualitative} presents qualitative results across multiple nuScenes sequences.
As shown in Figure \ref{fig:qualitative}a, the baseline model without semantic supervision struggles to detect distant vehicles and has difficulty discerning individual instances in cluttered, close-range scenarios during nighttime conditions.
This misidentification effect is further evident in Figure \ref{fig:qualitative}b, where TGRIP uses CLIP-based semantic cues to maintain more robust instance identities than the unsupervised baseline.
Finally, Figure \ref{fig:qualitative}c illustrates TGRIP's superior performance in complex intersection situations, demonstrating enhanced long-range spatial awareness and stable identification, both of which are essential for safe motion planning in autonomous driving.

Figure \ref{fig:qualitative2} validates the distillation of semantic knowledge into our auxiliary head. A principal component analysis (PCA) projection of the BEV features reveals separable clusters for distinct vehicle categories (e.g., cars vs. trucks), indicating that the model effectively encodes categorical semantics.
Additionally, cosine similarity heatmaps reveal that peak activations occur precisely at the corresponding object locations when comparing these features and target text embeddings.
This demonstrates successful cross-modal alignment, proving that the model can retrieve specific instances using high-level semantic concepts.

%% file: sections/5_conclusion.tex
\section{Conclusion and Future Works}

In this paper, we introduce TGRIP, a novel framework for BEV instance prediction that leverages the rich latent space of vision-language models.
We enforce cross-modal alignment within the network's shared feature backbone by distilling instance-specific visual embeddings (e.g., from CLIP) via an auxiliary semantic head.
This joint supervision strategy is crucial as it improves the representation of the core BEV feature during training without adding computational overhead during prediction inference and enhances the performance of nuScenes compared to the unsupervised baseline, establishing a new SOTA performance.
Furthermore, our ablation and qualitative analyses confirm that fine-grained visual cues effectively guide the network in learning robust, categorical latent representations. This semantic enrichment ultimately significantly improves instance disambiguation and long-range spatial awareness, providing a critical foundation for safer motion planning in autonomous driving systems.

Despite the promising results, several directions remain open for future investigation:
\begin{itemize}

    \item \textbf{Generalization across datasets and sensor modalities.}
     The current evaluation is conducted exclusively on nuScenes, which covers only a limited range of driving environments and uses a fixed set of sensors.
     A natural next step would be to validate TGRIP on datasets with different characteristics.
     Integrating LiDAR or radar modalities into the BEV feature extraction pipeline, beyond camera-based settings, could provide complementary geometric cues that would further benefit semantic alignment.
     This would also assess the effectiveness of the CLIP distillation strategy under richer or sparser sensor inputs.

    \item \textbf{Knowledge distillation into lightweight deployment models.}
    The inference efficiency of TGRIP is achieved by removing the semantic head at test time even although the backbone itself is designed for research-grade hardware.
    One important area for future research is to study whether the enriched BEV representations learned under CLIP supervision can be transferred to smaller, more efficient architectures through knowledge distillation.
    Specifically, a compact student model could be trained to mimic the intermediate BEV features of the TGRIP backbone while inheriting the semantic structure imposed during training, thereby addressing the latency and memory constraints of embedded automotive processors.
    This would bridge the gap between the improved representational quality demonstrated in this study and the requirements for real-world deployment on in-vehicle hardware.

    \item \textbf{Moving beyond dataset-bound supervision with open-vocabulary embeddings.}
    A fundamental limitation of the proposed pipeline is that it depends on 3D bounding box annotations in order to generate the CLIP-based semantic ground truth.
    This ties the supervision signal to the closed set of categories present in the training dataset, requiring accurate per-instance labels at every timestep.
    One potential solution to this issue is to explore open-vocabulary supervision strategies, where semantic embeddings are derived directly from raw image regions, rather than relying on predefined labels.
    TGRIP could leverage unannotated data and generalise to novel object categories encountered in long-tail driving scenarios with approaches based on region-level vision-language models or self-supervised patch descriptors.

    \item \textbf{BEV features as semantic-geometric input for downstream driving models.}
    The BEV representations produced by TGRIP are shaped by a combination of instance motion, spatial geometry and high-level visual semantics inherited from CLIP.
    This makes them an especially expressive input modality for downstream decision-making systems.
    One interesting direction to explore is the use of these features as a structured world representation for large language model-based or vision-language-action (VLA) driving systems, where a thorough understanding of the scene is crucial for creating interpretable and generalisable driving policies.
    Instead of relying on raw sensor data or manually crafted scene descriptors, such models could benefit from the compact, semantically organised BEV tokens produced by TGRIP.
    This could potentially improve the quality and explainability of the resulting driving behaviour.

\end{itemize}